\documentclass[11pt]{article}

\usepackage[preprint]{acl}

\usepackage{times}
\usepackage{latexsym}
\usepackage{multirow}
\usepackage{amsmath}
\usepackage{amssymb}
\usepackage{amsfonts}
\usepackage{stmaryrd}
\usepackage[T1]{fontenc}

\usepackage[utf8]{inputenc}

\usepackage{microtype}

\usepackage{inconsolata}

\usepackage{graphicx}
\usepackage{booktabs}

\title{Multilingual Coreference Resolution via\\ Cycle-Consistent Machine Translation}

\author{
\textbf{Adriana-Valentina Costache}$^{*}$\textbf{, Eduard Poesina}$^{*}$\textbf{, Silviu-Florin Gheorghe,}\\
\textbf{Paul Irofti, Radu Tudor Ionescu}$^{\diamond}$\\
Department of Computer Science, University of Bucharest, Romania\\
$^{*}$Equal contribution. $^{\diamond}${\texttt{raducu.ionescu@gmail.com}}
}

\begin{document}
\maketitle
\begin{abstract}
Coreference resolution is a core NLP task, having a broad range of downstream applications, e.g.~machine translation, question answering, document summarization, etc. While the task is well-studied in English, comparatively less attention is dedicated to coreference resolution in other languages, especially low-resource ones. To mitigate this gap, we propose a novel coreference resolution pipeline that harnesses machine translation (MT) from English to a target low-resource language, to generate or expand training data. To automatically validate the quality of the translated samples, we back-translate the samples and assess the similarity with the original English samples via cosine similarity in the latent space of a BERT model. The resulting similarity scores are integrated into the loss function to weight training samples according to their MT cycle consistency. Extensive experiments on four low-resource languages show that our pipeline brings significant performance gains in coreference resolution. Moreover, our pipeline enables accurate coreference resolution in languages where no previous corpora were available.
\end{abstract}

\setlength{\abovedisplayskip}{3.5pt}
\setlength{\belowdisplayskip}{3.5pt}
\setlength{\abovedisplayshortskip}{3pt}
\setlength{\belowdisplayshortskip}{3pt}

\vspace{-0.1cm}
\section{Introduction}
\vspace{-0.1cm}

Coreference resolution (CR) is a fundamental NLP task, which aims to identify all expressions in a text that refer to the same entity. The first attempts at solving the CR problem were heavily based on human-designed rules for the English language \cite{hobbs1978resolving, Ng2005SupervisedRF, ponzetto_NAACL_2006, Raghunathan_EMNLP_2010}. These types of methods are limited by the difficulty of drawing a complete list of non-contradictory rules, and are exposed to problems associated with the statistical nature of language. The foundational work of \citet{LEE_EMNLP_2017} was set to address CR by creating a fully trainable solution, without human-designed linguistic rules. The authors introduced the first end-to-end neural system, using a bidirectional LSTM to produce contextual span representations for joint mention detection in English. Deep models later benefited from the emergence of better neural encoders \cite{joshi-etal-2019-bert}, such as BERT \cite{Devlin-NAACL-2019}. While end-to-end models reach competitive results \cite{kirstain_ACL_2021, xu_EMNLP_2020}, they usually have many task-specific hyperparameters and are hard to tune, as stated by \citet{zhang_EMNLP_2023}. 

More recently, researchers introduced a new category of sequence-to-sequence solutions \cite{urbizu_CRAC_2020,liu-EMNLP-2022,bohnet-TACL-2023,straka-2023-ufal}, aiming to generate text representations of entity clusters. Notably, CorPipe \cite{straka-2023-ufal} won the CRAC 2023 shared task on multilingual coreference resolution, while CorPipeEnsemble ranked first in the CRAC 2025 (unconstrained) edition. Another direction of study is the use of zero-shot large language models (LLMs) via prompting. \citet{le_arxiv_2023} found that, although the zero-shot performance of promoted LLMs is respectable, they still remain way below specialized state-of-the-art models, by 10-20\% on benchmarks like CoNLL-2012/OntoNotes \cite{pradhan2012conll2012}. The CRAC 2025 results \cite{novak-etal-2025-findings} also indicate that zero-shot LLMs lag far behind specialized models, with a clear gap of about 13\% in terms of F1. 

These empirical observations underline the utility of task-specific datasets used to train and test specialized CR models. However, CR datasets in certain languages are small, outdated, or entirely missing. There are clear efforts to remedy this situation, e.g.~the CRAC 2025 shared task \cite{novak-etal-2025-findings} describes CorefUD as a harmonized multilingual collection of 22 datasets in 17 languages. By contrast, for low-resource languages, such as Romanian, we did not find any CR datasets that can be used in the evaluation and training of specialized models. To make things worse, it is reasonable to expect that the zero-shot performance in such languages is even lower. 

\begin{figure}[t!]
\centering
    \includegraphics[width=1.0\linewidth]{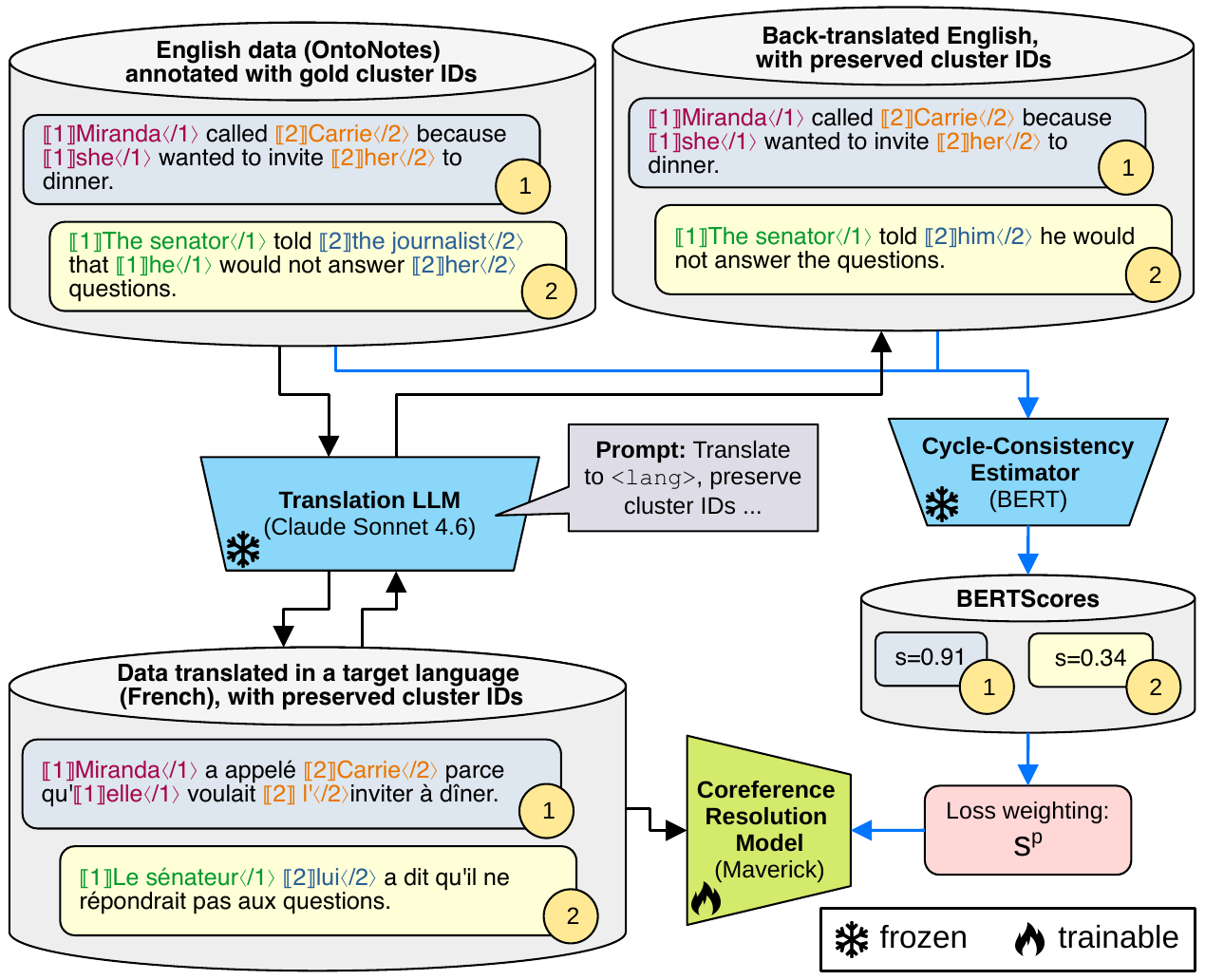}
    \vspace{-0.65cm}
\caption{Overview of the proposed pipeline for coreference resolution. An LLM, namely~Claude Sonnet 4.6 \cite{Anthropic-MC-2026}, is prompted to translate annotated samples from English to the target language and back. The cycle consistency of back-translations is estimated via BERTScore \cite{zhang2020bertscore}. Finally, the CR model, namely Maverick \cite{Martinelli-ACL-2024}, is trained on the target language, weighting the loss of each translated sample with $s^p$, where $s$ represents the BERTScore of the respective sample, and $p$ is a hyperparameter that controls the importance of cycle consistency. Best viewed in color.}
\label{fig:pipeline}
\vspace{-0.4cm}
\end{figure}

To this end, we propose a novel CR framework that leverages existing English resources via machine translation (MT) to generate new training data in a target low-resource language. As illustrated in Figure \ref{fig:pipeline}, we employ back-translation and assess the overlap between original and back-translated English samples, where the overlap is given by the cosine similarity computed in the embedding space of a pre-trained BERT model \cite{Devlin-NAACL-2019,zhang2020bertscore}. We conjecture that the utility of a translated data sample is proportional to its cycle consistency, i.e.~the cosine similarity of its back-translation. Therefore, we integrate the cosine similarity between original and back-translated English samples into the loss function, to weight the importance of translated samples according to their cycle consistency.

To validate the proposed framework, we perform experiments across four low-resource languages: French, Hungarian, Romanian and Russian. While three of these languages have small-scale publicly available CR datasets, there are no CR resources for Romanian. The results confirm that our cycle-consistent MT augmentation framework can significantly boost performance in CR across all four languages, in both training dataset expansion and training dataset generation scenarios.

In summary, our contribution is threefold:
\begin{itemize}
    \item \vspace{-0.18cm} We propose a novel CR framework based on MT to generate new training samples for low-resource languages, and modulate sample importance according to MT cycle consistency.
    \item \vspace{-0.22cm} We conduct comprehensive experiments across four low-resource languages, showing that the proposed framework can significantly boost CR performance.
    \item \vspace{-0.22cm} We manually curate a coreference resolution test set for Romanian, thus enabling the evaluation of CR systems for this low-resource language.
\end{itemize}

\vspace{-0.1cm}
\section{Method} 
\label{sec:methods}
\vspace{-0.1cm}

Our model extends Maverick~\cite{Martinelli-ACL-2024} with three modifications, to make it suitable for CR in low-resource languages. First, we replace the English-only encoder DeBERTa-v3-large~\cite{he2023debertav3} with mmBERT-base~\cite{marasovic2024mmbert}, a multilingual encoder pre-trained on 200+ languages. In this way, a single model can be employed across multiple languages. Second, we separate training into two phases: (i) train the mention detector with the frozen encoder, and (ii) fine-tune the encoder and the coreference heads using gold mentions as input, isolating the linking signal from mention-detection noise. Third, we augment the bilinear coreference scorer~\cite{LEE_EMNLP_2017} with MT cycle consistency, providing a discriminative signal, independent of encoder representations.

\noindent
\textbf{Generating data via MT.} The lack of large-scale CR resources in many languages motivates our MT-based augmentation strategy. We employ a highly capable LLM to perform MT, namely Claude Sonnet 4.6 \cite{Anthropic-MC-2026}. As illustrated in Figure ~\ref{fig:pipeline}, each source (English) document is translated using Claude Sonnet \cite{Anthropic-MC-2026} via zero-shot prompting (the exact prompt is specified in Table \ref{tab:prompt}). The prompt instructs the model to produce a fluent target-language translation, while preserving every $\llbracket k \rrbracket \ldots \langle/k\rangle$ span around the target-language equivalent of each English mention, thus maintaining all cluster identifiers $k \in \{1,2,...,K\}$, where $K$ is the number of entities. 

\noindent
\textbf{Back-translation quality scoring.} Translation errors introduce noise into the projected annotations. To quantify this noise per document, we employ back-translation to complete the translation cycle: each target language translation is itself submitted to Claude Sonnet 4.6 with a symmetric prompt requesting translation back to English, while preserving all cluster markers. The back-translated English text is then compared with the original English source via BERTScore \cite{zhang2020bertscore}, yielding a per-document quality score $s \in [0, 1]$. Our intuition is that a high-fidelity translation followed by a faithful back-translation recovers text semantically close to the source, whereas a translation that drops or misaligns mentions produces a divergent back-translation.

\noindent
\textbf{Per-document loss weighting.} Rather than applying a hard threshold to discard low-quality documents, which would lose potentially useful training data, we incorporate BERTScore directly into the training objective. Each document $D$ contributes with the following weight:
\begin{equation}\label{eq_weight}
w_D = s_D^{\,p},    
\end{equation}
where $s_D$ is the BERTScore between source and back-translated versions of document $D$, and $p \geq 0$ controls the strength of the penalty. Then, the weighted training objective becomes:
  \begin{equation}\label{eq_loss}
    \mathcal{L}(\theta) = \frac{1}{|\mathcal{D}|} \sum_{D \in \mathcal{D}} w_D \cdot \mathcal{L}_D(\theta),
  \end{equation}
  where $\mathcal{D}$ is the collection of translated documents (originally available in English), $\theta$ represents the parameters of the CR model, and $\mathcal{L}_D$ is the per-document loss for the current training phase. For training phase (i), $\mathcal{L}_D$ is the standard binary cross-entropy on mention start/end logits, while for phase (ii), it is the marginal log-likelihood over gold antecedents \cite{LEE_EMNLP_2017}.  
  

\begin{table*}[t]
    \centering
    \setlength{\tabcolsep}{5pt}
    \resizebox{0.85\linewidth}{!}{%
    \begin{tabular}{lc|c|ccc|ccc|ccc|ccc}
      \toprule
      \multicolumn{2}{r|}{\textbf{Language}$\rightarrow$} & \multicolumn{1}{c|}{\textbf{English}}
      & \multicolumn{3}{c|}{\textbf{French}}
      & \multicolumn{3}{c|}{\textbf{Hungarian}}
      & \multicolumn{3}{c|}{\textbf{Romanian}} 
      & \multicolumn{3}{c}{\textbf{Russian}} \\
      \multicolumn{2}{r|}{\textbf{Method}$\rightarrow$} 
      & Base    
      & Base & +MT & +$s^p$ 
      & Base & +MT & +$s^p$
      & ZS & +MT & +$s^p$
      & Base & +MT & +$s^p$ \\

      \midrule

      \multirow{3}{*}{\textbf{MUC}} & \textbf{P}
      & 95.0
      &85.2 &85.5 & \textbf{89.3}
      &87.3 & 87.5 & \textbf{88.9}
      & 70.1 &86.2 & \textbf{87.8}
      &95.5 &95.6 & \textbf{95.9} \\

       & \textbf{R}
      & 95.7
      &82.3 &83.4 & \textbf{88.0}
      &88.9 & 89.5 & \textbf{90.8}
      & 66.0 &84.8 & \textbf{86.5}
      &95.8 &96.2 & \textbf{96.5} \\

       & \textbf{F1}
      & 95.4
      &83.7 &84.4 & \textbf{88.6}
      &88.1 & 88.5 & \textbf{89.8}
      & 68.0 &85.5 & \textbf{87.1}
      &91.7 &92.0 & \textbf{92.4} \\
\midrule
      
      \multirow{3}{*}{\textbf{B$^{3}$}} & \textbf{P}
      & 88.1
      &80.8 &80.2 & \textbf{85.6}
      &84.2 & 84.2 & \textbf{85.4}
      & 65.8 &81.5 & \textbf{83.2}
      &92.1 &92.0 & \textbf{92.3} \\

       & \textbf{R}
      & 91.7
      &79.0 &80.6 & \textbf{85.4}
      &89.3 & 90.2 & \textbf{91.2}
      & 62.4 &80.9 & \textbf{82.4}
      &93.5 &94.0 & \textbf{94.3} \\

       & \textbf{F1}
      & 89.9
      &79.9 &80.7 & \textbf{85.5}
      & 86.7 & 87.1 & \textbf{88.2}
      & 64.1 &81.2 & \textbf{82.8}
      &92.8 &92.9 &\textbf{93.3}  \\
\midrule

      \multirow{3}{*}{\textbf{CEAF-E}} & \textbf{P}
      & 92.7
      &78.1 &78.6 & \textbf{83.7}
      &87.8 & 88.2 & \textbf{89.1}
      & 62.8 &80.1 & \textbf{81.9}
      &92.6 &92.8 & \textbf{93.1}\\

       & \textbf{R}
      & 86.2
      &73.2 &73.5 & \textbf{79.0}
      &80.1 & 80.7 & \textbf{82.0}
      & 57.9 &75.4 & \textbf{71.1}
      &90.8 &91.1 & \textbf{91.6} \\

       & \textbf{F1}
      & 89.4
      &75.6 &76.0 & \textbf{81.3}
      &83.8 & 84.2 & \textbf{85.4}
      & 60.2 &77.7 & \textbf{79.4}
      &91.7 &92.0 & \textbf{92.4} \\
\midrule

      \multirow{1}{*}{\textbf{CoNLL}} 


       & \textbf{F1}
      & 91.6
      &79.7 &80.4 & \textbf{85.1}
      &86.2 & 86.5 & \textbf{87.8}
      & 64.1 &81.5 & \textbf{83.1}
      &93.4 &93.6 & \textbf{94.0} \\
      
      \bottomrule
    \end{tabular}%
    }
    \vspace{-0.25cm}
    \caption{Coreference resolution results on four target languages (French, Hungarian, Romanian, Russian), measured with the
   official CoNLL-2012 scorer. Best score per language is highlighted in \textbf{bold}. Legend: \textbf{base} -- Maverick trained on original target language data; \textbf{ZS} -- zero-shot LLM (when no original training data is available); \textbf{+MT} -- Maverick trained with translated examples; \textbf{+$\mathbf{s}^\mathbf{p}$} -- Maverick trained with translated examples and cycle-consistent loss weighting. For reference, we report results on English with the \textbf{base} model.}
    \label{tab:main_results}
    \vspace{-0.25cm}
  \end{table*}

\vspace{-0.1cm}
\section{Experiments}
\vspace{-0.1cm}

\noindent
\textbf{Datasets.} For French, we use the ANCOR corpus~\cite{muzerelle2014ancor}, which contains 530 transcripts of spontaneous spoken French drawn from interviews, conversations, and oral surveys. 
For Hungarian, we use SzegedKoref~\cite{vincze2018szegedkoref}, a dataset comprising 320 short editorial and news documents annotated for nominal coreference. For Russian, we use RuCor~\cite{toldova2014rucor}, a corpus formed of 180 texts covering news, scientific articles, blog posts and fiction. For French, Hungarian and Russian, where the native gold data is small or domain-restricted, we supplement the within-language training data with LLM-translated OntoNotes~5.0 documents to expand both volume and domain diversity. For data augmentation via MT, we choose OntoNotes~5.0~\cite{weischedel2013ontonotes,pradhan2012conll2012} as the source English corpus, as it spans a broad range of genres: newswire, broadcast news, broadcast conversation, magazine, web text, telephone speech, and biblical text. 
For Romanian, no publicly available coreference dataset exists. We therefore construct a Romanian dataset entirely from English documents drawn from OntoNotes~5.0~\cite{weischedel2013ontonotes,pradhan2012conll2012}. The documents are translated with Claude Sonnet 4.6, which is instructed to preserve annotations. Further, the corresponding Romanian test set is manually verified and corrected by a native Romanian speaker to ensure that translation, mention boundaries, and coreference links are correct. 

\noindent
\textbf{Evaluation measures.}
Following \citet{Martinelli-ACL-2024}, we employ three evaluation metrics, namely MUC \cite{vilain-etal-1995-model}, B$^3$ \cite{Bagga-ACL-1998}, and CEAF-E ($\phi_4$-CEAF) \cite{luo-2005-coreference}. For each of them, we report the precision (P), recall (R), and F1 scores. We also report the CoNLL F1 score, which is defined as the average F1 score of the MUC, B$^3$ and CEAF-E metrics.

\noindent
\textbf{Hyperparameter setup.} We train Maverick \cite{Martinelli-ACL-2024} using AdamW, with a learning rate of $10^{-4}$ and a mini-batch size of $16$. We use a gradient clipping of $1.0$. We train using early stopping with a patience of $20$ epochs, and select the best model via CoNLL F1 on validation data. All other hyperparameters are left to their default values. Our pipeline introduces a single extra hyperparameter into Maverick \cite{Martinelli-ACL-2024}, namely the power $p$ in Eq.~\eqref{eq_weight}. We tune $p$ on validation data for one language (French), considering values for $p \in \{0.5, 1, 2, 3\}$. The optimal value $p=3$ is kept across all languages to avoid overfitting in hyperparameter space.

\begin{figure}[t!]
\centering
    \includegraphics[width=1.0\linewidth]{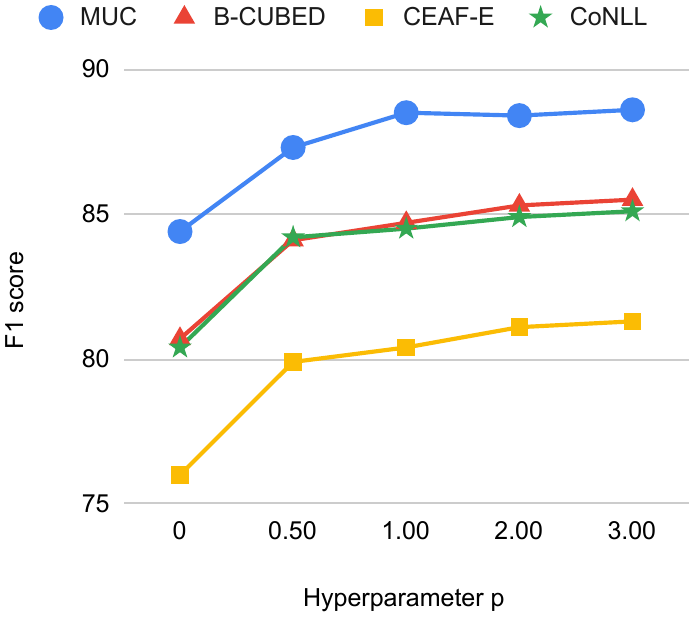}
    \vspace{-0.65cm}
\caption{Ablation of hyperparameter $p$, which controls the impact of loss weighting in Eq.~\eqref{eq_loss}. Best viewed in color.}
\label{fig:ablation_p}
\vspace{-0.5cm}
\end{figure}

\noindent
\textbf{Results.} For French, Hungarian and Russian, we compare three alternatives: a base Maverick model (trained on original within-language data), a Maverick model that benefits from MT data (trained on both original and translated data), and a Maverick model that benefits from MT data, but penalizes MT samples according to their cycle consistency (back-translation quality). For Romanian, there is no within-language data available, so we replace the base Maverick model with a zero-shot LLM (we use the same LLM as for MT, namely Claude Sonnet 4.6). In Table~\ref{tab:main_results}, we present comparative results across four low-resource languages. The results indicate that MT-based data augmentation is beneficial, especially for Romanian, where there are no available coreference resolution corpora. Furthermore, we observe additional performance gains when introducing cycle-consistent loss weighting. Here, the improvements stem primarily from higher precision on MUC and B$^3$, suggesting that loss weighting based on $s^p$ helps suppress spurious mentions introduced by translation artifacts.

\noindent
\textbf{Ablation of loss weighting hyperparameter.} In Figure \ref{fig:ablation_p}, we vary the hyperparameter $p$ in Eq.~\eqref{eq_weight}, considering values in the set $\{0, 0.5, 1, 2, 3\}$. Note that $p=0$ turns off the BERTScore weighting in Eq.~\eqref{eq_loss}. The ablation of $p$ is performed on the French dataset. The results show that higher values of $p$ lead to better results, confirming that our cycle-consistent loss weighting is very useful. However, going from $p=2$ to $p=3$, we observe that the performance gains begin to saturate.

\begin{table}[t]
\centering
\setlength{\tabcolsep}{4pt}
\resizebox{1.0\linewidth}{!}{%
\begin{tabular}{l|cccc}
\toprule
\textbf{Method} & \textbf{French} & \textbf{Hungarian} & \textbf{Romanian} & \textbf{Russian} \\
\midrule
Base & 79.7 & 86.2 & 64.1 & 93.4 \\
+$s^p$ (BLEU)       & 83.7 & 86.8 & 81.5 & 93.7 \\
+$s^p$ (BERTScore)  & \textbf{85.1} & \textbf{87.8} & \textbf{83.1} & \textbf{94.0} \\
\bottomrule
\end{tabular}
}
 \vspace{-0.25cm}
\caption{BLEU vs.~BERTScore comparison (in terms of CoNLL F1), as alternatives for the semantic similarity score $s$ used in our cycle-consistent loss weighting, across all four target languages.}
\label{tab:bleu_vs_bertscore}
 \vspace{-0.3cm}
\end{table}

\noindent
\textbf{BERTScore vs.~BLEU.}
In Table~\ref{tab:bleu_vs_bertscore}, we compare two alternatives to measure MT cycle consistency, namely BLEU~\cite{papineni-etal-2002-bleu} and BERTScore~\cite{zhang2020bertscore}. While both BLEU and BERTScore bring visible performance boosts across all four languages, BERTScore consistently outperforms BLEU. This is likely due to the fact that BLEU does not always capture semantic relations, such as synonymity. 

\vspace{-0.1cm}
\section{Conclusion}
\vspace{-0.1cm}

We proposed a novel pipeline for coreference resolution in low-resource languages, which harnesses MT to augment existing datasets or generate new training data (for languages where CR resources were not previously available). We assessed MT cycle consistency and introduced it in the loss function of the CR model to modulate the importance of translated data samples accordingly. To validate our approach, we conducted CR experiments across five low-resource languages. Our results demonstrated that our pipeline leads to significant performance gains, and even enables CR in languages without existing resources. In future work, we aim to expand the list of low-resource languages.

\section*{Acknowledgments}
This research is supported by the project ``Romanian Hub for Artificial Intelligence - HRIA'', Smart Growth, Digitization and Financial Instruments Program, 2021-2027, MySMIS no.~351416. This work is also supported by a grant of the Ministry of Research, Innovation and Digitization, CCCDI - UEFISCDI, project number PN-IV-P6-6.3-SOL-2024-0090, within PNCDI IV.

\section{Limitations}

Our framework leverages the use of a highly capable LLM in the translation phase. While translated data is central to our framework, as it brings significant performance gains, LLM usage can also represent a downside of our framework, introducing some limitations, as detailed below. 

First, LLMs are typically power-hungry models, having potentially negative effects on the environment due to their high energy consumption. As humanity will gradually move towards green energy production alternatives, the importance of the energy consumption problem of LLMs will diminish in the future. Moreover, we highlight that the translated data is meant to be reused multiple times to train and validate lighter models for coreference resolution. Hence, we limit LLM usage to the MT step, and refrain from fine-tuning LLMs for coreference resolution.

Second, potential biases of the LLM may eventually be transferred into the translated data, and later be inherited by the smaller coreference resolution model. We have manually inspected the translated examples and did not observe any age, gender, racial, or other kinds of biases. Nevertheless, our careful inspection does not completely exclude this possibility, especially for document genres and languages that are not included in our study.

\bibliography{custom}

\appendix

\section{Appendix}
\label{sec:appendix}

\vspace{-0.1cm}
\subsection{Translation Prompt}
\label{sec_hyper}
\vspace{-0.1cm}

To translate English documents annotated with entity clusters for coreference resolution, we employ Claude Sonnet 4.6 \cite{Anthropic-MC-2026}. The generic prompt used during translation from English to a target language \texttt{<lang>} is given in Table \ref{tab:prompt}. In the prompt template, \texttt{<lang>} is replaced with one of target languages, namely French, Hungarian, Romanian and Russian. To translate documents back to English, we use a symmetric prompt. The employed prompt comprises precise rules, especially regarding the preservation of annotations, which are particularly important for the underlying CR task. We also exemplify the rules via an example, to further explain to the LLM how the translation should be performed.

\begin{table*}[!t]
\centering
\small
\begin{tabular}{|p{0.93\linewidth}|}
\hline
\vspace{1.5mm}
{\ttfamily\small\raggedright
You are translating English text to \texttt{<lang>} while PRESERVING
coreference cluster annotations.

\medskip
The input text contains inline coreference markers:\\
\phantom{xx}{-}{-} \texttt{[[N]]word[/N]} marks a mention belonging to cluster N (an integer)\\
\phantom{xx}{-}{-} All mentions of the SAME entity share the SAME cluster ID\\
\phantom{xx}{-}{-} Markers can be nested: \texttt{[[1]]the CEO of [[2]]Acme[/2][/1]}

\medskip
YOUR TASK:\\
1.\ Translate the entire text to fluent, natural \texttt{<lang>}.\\
2.\ CRITICAL: every English mention \texttt{[[N]]...[/N]} MUST appear in the \texttt{<lang>}
translation with the SAME cluster ID N, wrapping the \texttt{<lang>}
equivalent of that mention.\\
3.\ Pronouns count as mentions. If ``he'' appears with ID 5 in English, the
\texttt{<lang>} equivalent pronoun (or whichever inflected form fits)
MUST also be marked \texttt{[[5]]...[/5]}.\\
4.\ If a mention is dropped because \texttt{<lang>} doesn't express it overtly
(e.g.\ pro-drop subject), still emit empty markers \texttt{[[N]][/N]} at the
dropped position to preserve the cluster.\\
5.\ Keep brackets balanced and properly nested.
Every \texttt{[[N]]} MUST have a matching \texttt{[/N]}.\\
6.\ Do NOT introduce new cluster IDs.
Use only the IDs present in the English text.\\
7.\ Output ONLY the \texttt{<lang>} text with annotations. No explanations,
no preamble, no markdown fences, just the annotated translation.

\medskip
EXAMPLE INPUT:\\
\texttt{[[1]]John[/1] went to [[2]]the store[/2]. [[1]]He[/1] bought [[3]]milk[/3], and}\\
\texttt{[[1]]John[/1] told [[4]]his wife[/4] about [[3]]it[/3].}

\smallskip
EXAMPLE OUTPUT:\\
\texttt{[[1]]Ion[/1] s-a dus [[2]]la magazin[/2]. [[1]]El[/1] a cump\u{a}rat [[3]]lapte[/3],}\\
\texttt{iar [[1]]Ion[/1] i-a spus [[4]]so\c{t}iei sale[/4] despre [[3]]asta[/3].}

\medskip
NOW TRANSLATE THIS TEXT (output the \texttt{<lang>} translation with annotations only):\\
\texttt{\{english\_text\}}
}
\vspace{1.5mm}
\\
\hline
\end{tabular}
\caption{Prompt used for Claude Sonnet 4.6 \cite{Anthropic-MC-2026} to translate
documents annotated with coreference resolution clusters from English to a
low-resource language \texttt{<lang>}, where \texttt{<lang>} is one of the
following four languages: French, Hungarian, Romanian, Russian. The output example is shown for Romanian.}
\label{tab:prompt}
\end{table*}

\subsection{Compute Environment}

We perform all our experiments on an academic compute environment, namely a workstation with a single Nvidia GeForce GTX 3090 GPU with 24 GB of VRAM. The reported results represent averages over three runs.

\subsection{Romanian Data Annotation}

The annotator employed to verify and correct the English$\rightarrow$Romanian translations is an adult who holds a master degree at a university located in Romania. The recruited annotator willingly agreed to engage in the annotation process, after agreeing to our terms and conditions. The authors provided accurate and complete instructions regarding the annotation task. The annotator was also given the LLM prompt. A fair compensation (25 EUR per hour) was paid to the annotator, upon completing the annotations. This is almost double the average wage in Romania (13.6 EUR per hour)\footnote{\url{https://www.romania-insider.com/eurostat-romanians-working-hours-salaries-april-2026}}. The authors verified the manual annotations to confirm that the annotation task was carefully completed by the recruited annotator, according to the provided instructions.

\subsection{Romanian Data License Agreement}

The Romanian version of OntoNotes~5.0 will be released under the LDC User Agreement for Non-Members.

\end{document}